\pdfoutput=1

\documentclass[11pt]{article}

\usepackage[final]{acl}

\usepackage{times}
\usepackage{latexsym}

\usepackage[T1]{fontenc}

\usepackage[utf8]{inputenc}

\usepackage{microtype}

\usepackage{inconsolata}

\usepackage{graphicx}
\usepackage{booktabs}
\usepackage{amsmath}

\usepackage[dvipsnames]{xcolor}
\usepackage{fontawesome5}
\usepackage{subcaption}
\usepackage{tcolorbox}
\tcbuselibrary{breakable}
\usepackage{caption}
\usepackage{multirow} 
\usepackage{threeparttable}
\usepackage{float}

\title{\textsc{Evade}: LLM-Based Explanation Generation and Validation \\ for Error Detection in NLI}

\author{
\textbf{Longfei Zuo\textsuperscript{1, 2}\thanks{Main work carried out while at LMU Munich.}} \quad
\textbf{Barbara Plank\textsuperscript{2, 3}} \quad
\textbf{Siyao Peng\textsuperscript{2, 3}} \quad
\\
\textsuperscript{1} Technical University of Munich, Heilbronn, Germany \\
\textsuperscript{2} MaiNLP, Center for Information and Language Processing, LMU Munich, Germany \\
\textsuperscript{3} Munich Center for Machine Learning (MCML), Munich, Germany \\
\href{mailto:longfei.zuo@tum.de}{\texttt{\textcolor{black}{longfei.zuo@tum.de}}} \quad
\texttt{\textcolor{black}{
\href{mailto:b.plank@lmu.de}{\textcolor{black}{b.plank@lmu.de}} \quad
\href{mailto:loganpeng1992@gmail.com}{\textcolor{black}{loganpeng1992@gmail.com}}}}
}

\newcommand{\framework}{\textsc{Evade}}
\newcommand{\varierr}{\textsc{VariErr}}

\begin{document}

\maketitle

\begin{abstract}
High-quality datasets are critical for training and evaluating reliable NLP models.
In tasks like natural language inference (NLI), human label variation (HLV) arises when multiple labels are valid for the same instance, making it difficult to separate annotation errors from plausible variation.
An earlier framework, \varierr{} \citep{weber-genzel-etal-2024-varierr}, asks multiple annotators to explain their label decisions in the first round and flags errors through validity judgments in the second round.
However, conducting two rounds of manual annotation is costly and may limit the coverage of plausible labels or explanations.
Our study proposes a new framework, \framework{}, for generating and validating explanations to detect errors using large language models (LLMs).
We perform a comprehensive analysis comparing human- and LLM-detected errors for NLI across distribution comparison, validation overlap, and impact on model fine-tuning.
Our experiments demonstrate that LLM validation refines generated explanation distributions to more closely align with human annotations, and that removing LLM-detected errors from training data yields improvements in fine-tuning performance than removing errors identified by human annotators.
This highlights the potential to scale error detection, reducing human effort while improving dataset quality under label variation.

\end{abstract}

\section{Introduction}

\begin{figure}[t]
    \centering 
    \includegraphics[width=\columnwidth]{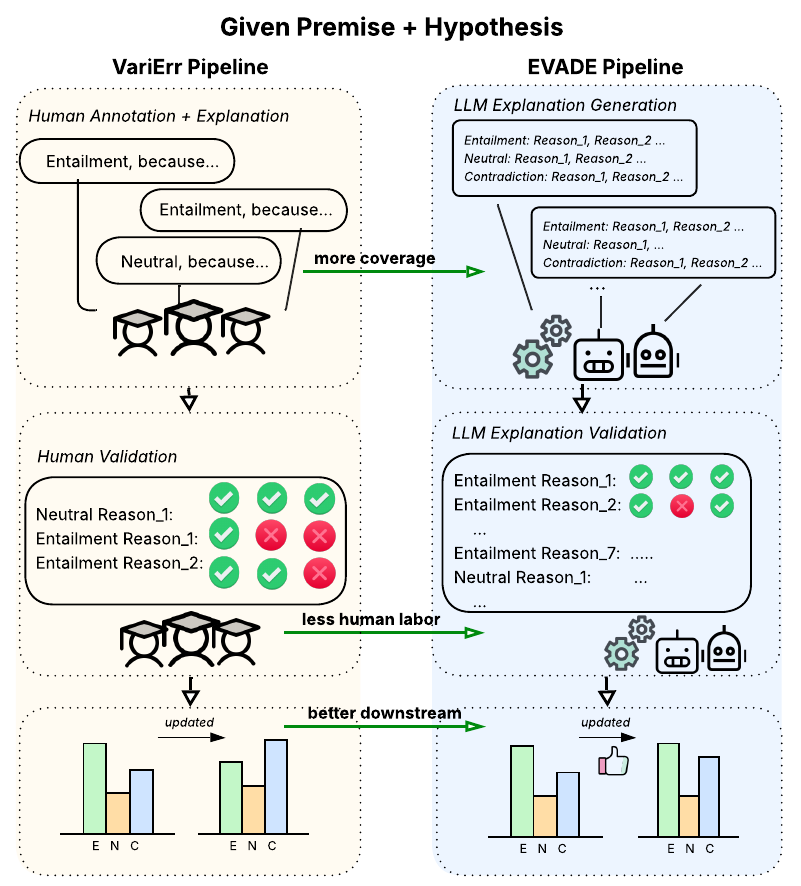}
    \caption{Overview of our LLM-based \textsc{Evade} framework compared with the human-based \varierr{} pipeline \citep{weber-genzel-etal-2024-varierr}. The first two modules, explanation generation and validation, are the core components. Compared with \varierr{}, our \framework{} framework provides broader explanation coverage, requires less human intervention, and delivers better downstream performance in predicting label distributions.}
    \label{fig:pipeline}
\end{figure}

Datasets are fundamental to research, yet annotation errors can undermine model reliability \cite{aqua}, highlighting the need for high-quality data in trustworthy Natural Language Processing (NLP) systems \cite{klie2024analyzing,10534765,weber-genzel-etal-2024-varierr}.
Human Label Variation (HLV, \citealt{plank-2022-problem}) refers to plausible variation in annotations, in which multiple labels can be assigned to a single instance. 
HLV has gained significant attention, particularly for Natural Language Inference (NLI) \cite{pavlick-kwiatkowski-2019-inherent, nie-etal-2020-learn}, where multiple plausible labels (Entailment, Neutral, or Contradiction) can be valid for the same premise-hypothesis pair \cite{jiang-marneffe-2022-investigating, jiang-etal-2023-ecologically, jayaweera2025disagreement}. 
While HLV better reflects real-world ambiguities, it also introduces a challenge: annotation errors may be obscured by the variation.

Many approaches for annotation error detection (AED) have been introduced in previous research \cite{klie-etal-2023-annotation, weber-plank-2023-activeaed, weber-etal-2024-donkii,bernier2024annotation}, but to our knowledge, \citet{weber-genzel-etal-2024-varierr} is the only work that explicitly focuses on separating errors from HLV.
They propose a two-round error detection procedure in which expert annotators first provide labels with explanations and then reassess their reasoning after reviewing the full set of labels and explanations from all the annotators.
For each annotated NLI label, if the annotators do not validate their own previously written explanations in the second round, the label is considered erroneous. 
Previous studies typically introduce artificial noise by randomly flipping labels \cite{zheng2021meta,jinadu-ding-2024-noise}, but these synthetic errors are often easily identifiable \cite{larson-etal-2019-outlier}.
In contrast, by leveraging self-validation, \varierr{} captures naturally occurring errors and enables a more realistic evaluation of error detection. 

However, a key limitation of this approach is its reliance on human experts to generate and validate explanations, both resource-intensive and difficult to scale. 
Fortunately, recent work \cite{chen2025rosenamellmgeneratedexplanations} shows that LLM-generated explanations are comparable to humans in approximating label distributions in NLI, suggesting the potential to extend the VariErr framework to an LLM-based pipeline.

This paper presents \framework{}, an LLM-driven method for error detection that leverages explanation generation and validation.
Our main research questions are:
\textbf{RQ1:} How effectively can LLM-generated explanations reflect valid HLV, and how do these compare to human-written explanations?
\textbf{RQ2:} To what extent does LLM-validation enable reliable error detection compared to humans?
\textbf{RQ3:} What is the impact of removing LLM-detected errors on downstream fine-tuning performance in terms of alignment with HLV? \footnote{Our code and data are publicly available at \url{https://github.com/mainlp/LLM_AED}.}

Figure \ref{fig:pipeline} presents the structure of \framework{}.
We first prompt LLMs to generate explanations given a premise–hypothesis pair and a candidate label. 
Subsequently, LLMs evaluate the explanations by assigning validity scores, indicating how effectively each explanation supports the assigned label (\S\ref{sec:framework}).
To assess the effectiveness of our framework, we compare it with the human-driven pipeline from multiple perspectives, including explanation distribution, validated label overlap, linguistic similarity, and downstream task performance. 
Our findings demonstrate that \framework{} aligns closely with human annotators in terms of label agreement, and can effectively identify errors, yielding annotations that more accurately reflect HLV (\S\ref{sec:comparison} \& \S\ref{sec:downstream-finetuning}).

\section{Related Work}

\textbf{Human label variation (HLV)} refers to cases where annotators assign different but plausible labels to the same instance \cite{plank-2022-problem}.
HLV may arise from subjectivity \cite{10.1609/aaai.v37i6.25840}, semantic ambiguity in the target instances \cite{aroyo2013crowd, truthisalie} or guideline divergence \cite{peng-etal-2024-different}, which challenges the assumption of a single ground truth in annotation. \citet{aroyo2013crowd} propose the crowd truth, capturing subjective and diverse human interpretations.

HLV has been widely examined in the context of Natural Language Inference (NLI).
\citet{pavlick-kwiatkowski-2019-inherent} show that NLI annotation disagreements are often systematic rather than noise.
ChaosNLI \cite{nie-etal-2020-learn} provides a human judgment distribution (HJD, \citealt{chen-etal-2024-seeing}) by
crowd-sourcing 100 annotations per instance over a subset of MNLI \cite{williams-etal-2018-broad}, SNLI \cite{bowman-etal-2015-large}, and $\alpha$NLI \citep{Bhagavatula2020Abductive} instances.
LiveNLI \citep{jiang-etal-2023-ecologically} and VariErrNLI \citep{weber-genzel-etal-2024-varierr} supply ecologically valid explanations that verbalize different reasoning behind label decisions. 

\noindent
\textbf{Annotation error detection (AED)} is crucial in  
NLP, as many widely-used benchmarks contain annotation errors \cite{northcutt2021pervasivelabelerrorstest,klie-etal-2023-annotation,bernier2024annotation,larson-etal-2020-inconsistencies, weber-etal-2024-donkii}.
\citet{klie-etal-2023-annotation} conducted a comprehensive survey of error detection methods focusing on single-label tasks. However, annotation errors also arise when multiple labels are valid at the same time.
To separate annotation error from HLV, VariErrNLI \cite{weber-genzel-etal-2024-varierr} incorporates a second round of validity judgment to assess whether the provided explanations support the annotator's label decisions. 

\noindent
\textbf{LLM-generated explanations} are frequently used across a variety of tasks \citep{kunz-kuhlmann-2024-properties}, addressing whether they can approximate human reasoning.
Within NLI, recent studies investigate how LLM-generated explanations can be leveraged to understand and model human annotation variations.
\citet{jiang-etal-2023-ecologically} find that GPT-3 can generate fluent explanations, though not always aligned with the target label. 
\citet{chen2025rosenamellmgeneratedexplanations} prompt LLMs to generate explanations guided with human-annotated labels and achieve similar performance in approximating human judgment distribution as when using human-written explanations. 
\citet{hong2025litexlinguistictaxonomyexplanations} propose a linguistic taxonomy to further guide LLMs in generating more diverse explanations. 
The studies above highlight the potential of LLMs in generating explanations.
Our paper addresses LLMs' ability to validate their explanation generation and detect annotation errors.

\section{Generating and Validating Explanations Using LLMs}\label{sec:framework}

This section details our \framework{} pipeline using LLMs.
We introduce our dataset and model setups (\S\ref{subsec:framework-setup}), explanation generation and filtering process (\S\ref{subsec:framework-generation}), as well as validation scenarios (\S\ref{subsec:framework-validation}).

\subsection{Setups}\label{subsec:framework-setup}

\paragraph{Dataset}
We experiment on the \varierr{} dataset \citep{weber-genzel-etal-2024-varierr}, including NLI labels, explanations, and second-round validations from four human expert annotators. 
All  500 \varierr{} instances are also in ChaosNLI \citep{nie-etal-2020-learn}, containing label distributions from 100 crowdworkers.
This setup provides references for evaluating label distribution and explanation generation.

\paragraph{Models}
We select four state-of-the-art LLMs from two families due to their demonstrated capabilities in instruction following, critical in our generation and validation experiments: \textit{Llama-3.1-8B-Instruct}, \textit{Llama-3.3-70B-Instruct} \cite{grattafiori2024llama},\footnote{At submission time, v3.1 is the most recent Llama-8B-Instruct model release and v3.3 for Llama-70B-Instruct.} \textit{Qwen2.5-7B-Instruct} and \textit{Qwen2.5-72B-Instruct} \cite{qwen2.5}.

\subsection{Explanation Generation and Filtering}
\label{subsec:framework-generation}

\paragraph{Generation}
We follow \citet{chen2025rosenamellmgeneratedexplanations} to generate all distinct explanations for each label in a given premise-hypothesis pair to obtain a broad coverage.
To maintain output validity, we adapted their prompt to allow LLMs to abstain if the label cannot be reasonably justified.
Our prompt also explicitly prohibits introductory phrases or semantic repetitions.
We kept all qualified explanation generations in our repository; thus, \citet{chen2025rosenamellmgeneratedexplanations}'s label-guided or label-free approaches in picking the longest or first explanations are irrelevant to this study. See Appendix \ref{subsec:appx-generation-prompt} Figure \ref{fig:generate_explanation} for the template.

\paragraph{Filtering}

Despite explicit instructions, we still find two main issues with LLM-generated explanations: \textit{fallback responses} and \textit{formatting errors}, which would not occur in human-written explanations. 
First, rather than omitting the output when uncertain, 
Llama-72B model sometimes responds with fallback statements such as \textit{``Note: Since the statement is not supported by the context, there are no explanations for why the statement is true.''}
These responses signal the model’s uncertainty or disagreement with the target label. While they may be interpreted as a form of ``non-variation'', they fail to provide substantive explanatory content for any specific label.
Second, we observe formatting issues in the outputs of certain models, such as truncated outputs from Llama-8B caused by the default generation limit (256 tokens), and occasionally generated Chinese sentences from Qwen-7B, even all prompts are in English.
As our framework aims not only to evaluate error detection performance but also to compare the alignment of LLM-generated explanations and relevant label distributions with human annotations, we manually filter out such explanations to ensure a cleaner distribution. \footnote{In practice, we removed 37 incomplete outputs from Llama-8B, excluded 17 fallback generations from Llama-70B, and filtered out 6 explanations containing Chinese texts from Qwen-7B.}

\paragraph{Results}

\begin{table}[t]
\centering
\resizebox{\columnwidth}{!}{%
\begin{tabular}{l|cc|cc}
\toprule
\textbf{Model} & \textbf{\# Expl.}  & \textbf{Avg. Length} 
& \textbf{\# Label/Item} & \textbf{\# Expl./Label}
\\
\midrule
Llama-8B & \textbf{8845} & 17.60  
& 3.00  & \textbf{5.90} 
\\
Llama-70B & 4022 & \textbf{22.35}
& 3.00  & 2.68 
\\
Qwen-7B & 4047  & 16.19
& 2.87  & 2.82
\\
Qwen-72B & 3920  & 18.43
& 3.00  & 2.61 
\\
\midrule
VariErr &  1933 & 13.89
& 1.76  & 2.20 
\\
\bottomrule
\end{tabular}
}
\caption{Generation statistics on 500 VariErr examples. \# Expl.: total number of explanations; Avg. Length: average number of words per explanation; \# Label/Item: number of LLM-explained labels per instance; Expl./Label: average number of explanations per label.}
\label{tab:generation_stats}
\end{table}

We observe noticeable differences across model generation results in Table \ref{tab:generation_stats}. 
Llama-8B generates nearly twice as many explanations as the others, 
suggesting potential redundancy between explanations,
and Llama-70B yields the longest explanations (22.35 tokens on average) among all the models.
These discrepancies become more evident when compared with human annotations in \varierr{}. 
Human annotators generated much fewer explanations in quantity and, in general, much shorter explanations (13.89 tokens). 

We then look at the average number of NLI labels that are supported by LLM-generated explanations per instance (\#Label/Item), as well as the average number of explanations per label (\#Expl./Label). 
\varierr{} receives on average 1.76 explained labels.
However, LLMs tend to produce explanations for every label (or almost all in the case of Qwen-7B), which is unlikely to reflect a realistic human label variation.
As for the number of generated explanations per label, Llama-8B still produces twice as many as humans under each label, due to its extensive output.
These LLM over-generations necessitate the explanation validation step in the \framework{} framework.

\subsection{Explanation Validation}
\label{subsec:framework-validation}

To address the excessive generation of explanations, we instruct LLMs to validate their own explanations.
Precisely, we ask an LLM to give validity judgments on whether each of its generated explanations supports the associated label, mirroring \varierr{}'s second-round self-validation process. 
We aim to align LLMs' explanations and label distribution more closely with human annotations and to demonstrate that LLMs can likewise be used for explanation validation and error detection.

Following \citet{weber-genzel-etal-2024-varierr}, we prompt LLMs to assign a \textbf{validity score} (between 0.0 and 1.0) assessing how much each of its generated explanations justifies the corresponding label for the given premise–hypothesis pair.
We consider an explanation as \textbf{validated} if its validity score $p$ exceeds a predefined threshold $\tau$; otherwise, it is \textbf{not-validated}.
To determine whether a label is erroneous, we follow the criterion from \varierr{}: a label is regarded as an ``error'' if the scores of all associated explanations fall below the threshold $\tau$.

\paragraph{Prompts}
To systematically examine the role of context in LLM validation, we experiment with three prompting scenarios:
(1) \textbf{\textit{one-expl}}: an LLM scores its generated explanations one at a time, without reference to other explanations;
(2) \textbf{\textit{one-llm}}: an LLM scores each of its generated explanations in the context of all its generations; 
and (3) \textbf{\textit{all-llm}}: an LLM scores each of its generated explanations in the context of explanations generated by all four LLMs. 
When multiple explanations are in context, the LLM receives all explanations simultaneously and outputs a json file with set of scores in a single pass.
Our prompts are adapted from \citet{weber-genzel-etal-2024-varierr}; see Appendix \ref{subsec:appx-validation-prompt} Figures \ref{fig:validate-prompt-one-expl}-\ref{fig:validate-prompt-one-llm-all-llms} for details.

\paragraph{Results}

\begin{table}[t]
\resizebox{\columnwidth}{!}{%
\small
\centering
\begin{tabular}{lcccc}
\toprule
\textbf{Model} & \textbf{\textit{one-expl}} & \textbf{\textit{one-llm}} & \textbf{\textit{all-llm}} \\
\midrule
\multirow{1}{*}{Llama-8B}  & 0.8003\(_{\;0.1290}\)  & 0.5009\(_{\;0.3134}\)& 0.5154\(_{\;0.3616}\) \\
\midrule
\multirow{1}{*}{Llama-70B} & \textbf{0.8456} \(_{\;0.1126}\) & 0.6483 \(_{\;0.2114}\)& 0.6439 \(_{\;0.2333}\)\\
\midrule
\multirow{1}{*}{Qwen-7B}  & 0.7107 \(_{\;0.1442}\) & 0.5675 \(_{\;0.2628}\)& 0.3825\(_{\;0.3894}\) \\
\midrule
\multirow{1}{*}{Qwen-72B} & 0.8299 \(_{\;0.0780}\) & \textbf{0.7279} \(_{\;0.1352}\)& \textbf{0.7006} \(_{\;0.1835}\)\\
\bottomrule
\end{tabular}
}
\caption{Average validation scores and standard deviations across three prompting scenarios for four LLMs. }
\label{tab:avg_val_combined}
\end{table}

\begin{figure*}[t]
    \centering
    \includegraphics[width=\textwidth]{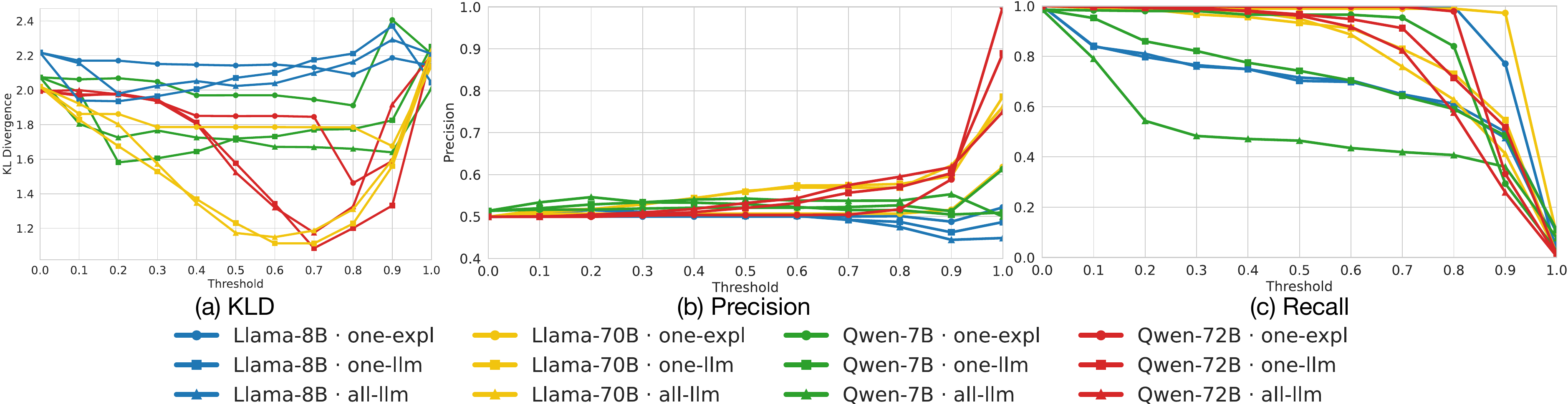}
    \caption{(a) shows the KL divergence curves between model distributions and ChaosNLI annotations across three prompting scenarios with validation threshold from 0.1 to 0.9. 
    (b) and (c) present the precision and recall of the LLM-validated labels, computed against the VariErr-validated labels as ground truth, with validation thresholds from 0.1 to 0.9.}
    \label{fig:threshold}
\end{figure*}

Table \ref{tab:avg_val_combined} presents the average validation scores of different LLMs across the three validation scenarios. 
The \textit{one-expl} setting consistently yields the highest validity, with most LLMs producing average validity scores above 0.8. This is the only prompt setup that does not provide contextual information from other candidate explanations, which most likely leads to models' over-confidence in the explanations' validity. 
When additional context is introduced in the \textit{one-llm} and \textit{all-llm} settings, models can review alternative explanations for the same instance. This broader perspective may result in more comprehensive evaluations.
However, providing contextual explanations generated by other LLMs (\textit{all-llm} versus \textit{one-llm}) only subtly changes the average validity score for all models except the Qwen-7B model. 
We then turn to the standard deviation (std), obtained by scoring each label for each instance and averaging across labels.
The \textit{one-expl} scenario achieves the lowest std, indicating more consistent scoring behavior for each label.
However, this stability may reduce sensitivity to differences between valid and invalid explanations within the same label. In contrast, \textit{one-llm} and \textit{all-llm} appear to balance multiple explanations with higher std, and can thereby filter relatively more invalid explanations and yield refined distributions.

Moreover, smaller models tend to assign lower scores than large ones across all settings.  
The gap between differently sized models becomes particularly large in the \textit{one-llm} and \textit{all-llm} settings.  
One possible explanation is that longer contextual inputs in these settings may overwhelm smaller models, reducing their confidence when confronted with multiple competing explanations.
For example, \textit{all-llm} provides an average of approximately 41.7 explanations per instance, showing an extreme difference in average validity score between Qwen-7B (0.3825) and Qwen-72B (0.7006).

However, one should be cautious in assuming that differences in average validity scores directly indicate that one model is better or worse.
To assess how well \framework{} aligns with humans,  we next examine whether the low validity scores correspond to genuine ``errors'' and analyze the overlap between LLM-validated labels and those approved by human annotators in \varierr{}.

\section{LLM versus Human Validation}
\label{sec:comparison}

After generation and validation, we compare the LLM-based (\framework{}) and human-based error detection (\varierr{} in \citealt{weber-genzel-etal-2024-varierr}) pipelines from different angles.
We analyze the label distribution associated with LLM-generated explanations before and after LLM validation (\S\ref{subsec:distribution-comparison}) and examine the validated label overlap with those from human annotators (\S\ref{subsec:overlap}). 
Since \varierr{}'s validity judgments are binary and our LLM outputs range from 0.0 to 1.0, we also show in these sections how much the validity threshold $\tau$ affects our comparison. 
We last assess the semantic similarity between LLM-generated and human-written explanations, both before and after validation (\S\ref{subsec:explanation-comparison}).

\subsection{Alignment with ChaosNLI Distribution}
\label{subsec:distribution-comparison}

Since HLV treats label distribution rather than a single label as the gold standard, we want to measure the similarity between label distributions derived from model-validated explanations with those from human annotations in ChaosNLI \citep{nie-etal-2020-learn}. The distributions in ChaosNLI are aggregated from 100 annotators, providing a softer representation that better reflects a reference ``gold'' distribution \citep{chen-etal-2024-seeing}.
Specifically, we examine model distributions before and after validation against the crowdworker distribution to assess whether the validation step improves the alignment of model outputs with human consensus.
We employ Kullback-Leibler divergence (KL, \citealt{kullback1951information}) following previous works \cite{chen-etal-2024-seeing, chen2025rosenamellmgeneratedexplanations} to measure the discrepancy between distributions.

An appropriate threshold has a substantial impact on the resulting distribution. To assess this effect, we compute the KL divergence between the model-validated distributions and the ChaosNLI annotations across three validation scenarios and a range of validation thresholds. Results are presented in Figure \ref{fig:threshold}a.
A threshold of \textbf{$\tau$=0.0} represents the pre-validation condition, as the validation procedure does not permit negative scores. 
After validation, KL divergence generally decreases across most models and prompting scenarios as the threshold increases up to approximately \textbf{$\tau$=0.8}.
This reduction is particularly pronounced for the larger models, indicating better agreement with the human label distribution. 
The point in the plot where the KL divergence is minimized corresponds to the validation threshold that yields the closest alignment with the ChaosNLI distribution. 
However, as the threshold increases further, the alignment with the human distribution weakens. 
This suggests that while a high validation threshold can filter out invalid information, an overly strict threshold may also inadvertently discard correct information.

\subsection{Comparing with Validated \varierr{}}
\label{subsec:overlap}

\begin{table*}[h]
\centering
\renewcommand{\arraystretch}{1.15}
\resizebox{\linewidth}{!}{
\begin{tabular}{lcccc}
\toprule
\multicolumn{1}{c}{\textbf{Models}} & \textbf{Lexical\ n=3 $\uparrow$} & \textbf{Syntactic\ n=3 $\uparrow$} & \textbf{Cosine$\uparrow$} & \textbf{Euclidean$\uparrow$} \\
\midrule
within-human & 0.052 & 0.144 & 0.529 & 0.522 \\
\midrule
\multicolumn{5}{l}{\textit{within-LLM} \,\,} \\
Llama-8B  
& \textbf{0.103} / 0.101 / \underline{0.097} / 0.099 
& \textbf{0.265} / 0.264 / \underline{0.263} / \underline{0.263} 
& \underline{0.599} / 0.608 / \textbf{0.611} / 0.606 
& \underline{0.542} / \textbf{0.545} / \textbf{0.545} / 0.544  \\
Llama-70B 
& 0.030 / \textbf{0.031} / \underline{0.029} / \underline{0.029} 
& 0.201 / 0.201 / \underline{0.200} / \textbf{0.202} 
& \underline{0.595} / 0.606 / 0.605 / \textbf{0.608} 
& \underline{0.535} / 0.539 / 0.538 / \textbf{0.540} \\
Qwen-7B   
& \underline{0.010} / \textbf{0.012} / \underline{0.010} / \underline{0.010} 
& \underline{0.160} / 0.161 / 0.164 / \textbf{0.170} 
& \underline{0.533} / \textbf{0.548} / 0.536 / 0.536 
& \underline{0.517} / \textbf{0.521} / 0.518 / 0.518  \\
Qwen-72B  
& \textbf{0.023} / \textbf{0.023} / \underline{0.021} / 0.022 
& 0.188 / 0.189 / \underline{0.187} / \textbf{0.191} 
& \underline{0.578} / 0.583 / 0.583 / \textbf{0.584} 
& \underline{0.530} / \textbf{0.532} / 0.531 / \textbf{0.532} \\
\midrule
\multicolumn{5}{l}{\textit{LLM-vs-human} \,\,} \\
Llama-8B  
& \textbf{0.020} / \textbf{0.020} / 0.019 / \underline{0.018} 
& \underline{0.116} / \underline{0.116} / \textbf{0.117} / \underline{0.116} 
& \underline{0.450} / 0.452 / \textbf{0.453} / \textbf{0.453} 
& \underline{0.497} / \textbf{0.498} / \textbf{0.498} / \textbf{0.498} \\
Llama-70B 
& \underline{0.028} / 0.029 / 0.029 / \textbf{0.030} 
& \underline{0.129} / 0.131 / \textbf{0.132} / \textbf{0.132} 
& \underline{0.515} / 0.521 / 0.521 / \textbf{0.523} 
& \underline{0.514} / 0.516 / 0.516 / \textbf{0.517} \\
Qwen-7B   
& 0.015 / \textbf{0.016} / 0.015 / \underline{0.014} 
& 0.101 / \textbf{0.103} / 0.100 / \underline{0.095} 
& \underline{0.469} / \textbf{0.477} / \underline{0.469} / 0.470 
& \underline{0.502} / \textbf{0.504} / \underline{0.502} / \underline{0.502}  \\
Qwen-72B  
& \underline{0.023} / \textbf{0.024} / \underline{0.023} / \textbf{0.024} 
& \underline{0.125} / \textbf{0.127} / 0.126 / 0.126 
& \underline{0.502} / 0.506 / \textbf{0.507} / 0.504 
& \underline{0.511} / \textbf{0.512} / \textbf{0.512} / \textbf{0.512} \\
\bottomrule
\end{tabular}
}
\caption{Linguistic similarity (lexical, syntactic, semantic) of explanations across three evaluation regimes: \textit{within-human}, \textit{within-LLM}, and \textit{LLM-vs-human}. Each cell lists the scores in order: before validation, after \textit{one-expl} validation, after \textit{one-llm} validation, and after \textit{all-llm} validation. We \underline{underline lowest score(s) in each cell} and \textbf{bold highest}.}
\label{tab:liguistic-similarity}
\end{table*}

Given that LLMs generate substantially more explanations than humans and tend to provide explanations for every label, it is expected that LLMs identify more errors during validation, as their outputs may initially include a large number of inconsistent arguments.
To obtain a more reliable comparison between LLM- and human-validation, we therefore focus on \textbf{the overlap of validated labels} rather than on the overlap of detected errors.

When comparing the overlaps, we observe strong similarity. The majority of human-validated labels are also confirmed by the LLM validation process. 
We thereby compute the precision and recall of LLM-validated labels by considering the VariErr labels as the gold standard, varying the validation threshold from 0.1 to 0.9.

Figures \ref{fig:threshold}b-\ref{fig:threshold}c present the results.
For precision, all models start around P=0.5 when $\tau$=0. Larger models exhibit a generally increasing trend as the threshold rises, whereas smaller models show more fluctuation and a noticeable decline when the threshold exceeds 0.6.
This pattern suggests that low thresholds effectively filter out low-confidence predictions and reduce false positives. Still, for smaller models, overly high thresholds begin to remove true positives as well, leading to a drop in precision.
Regarding recall, all models across different validation scenarios exhibit a consistent decreasing trend as the threshold increases, with a particularly sharp decline beginning around 0.7. This occurs because higher thresholds exclude more true-positive labels, thereby reducing the overall recall.
Among the three validation modes, \textit{all-llm} is the most conservative: it validates substantially fewer labels and yields lower recall than the other two modes.

Overall, excessively low validation thresholds introduce numerous unreliable labels, while overly high thresholds risk discarding valid ones. An optimal threshold should balance precision and recall. Considering these metrics together with the KL divergence, we select the threshold with low KL divergence that also corresponds to relatively high precision and recall values as our final validation threshold (detailed in Appendix \ref{appx:validation-threshold} Table~\ref{tab:validation_threshold}).
Interestingly, the selected thresholds closely align with the average validation scores reported in Table~\ref{tab:avg_val_combined} across most models and prompting settings, supporting the calibration of validation scores as meaningful indicators, with low scores generally corresponding to genuine annotation errors.
These thresholds are set to facilitate experiments and analyses in the following sections.

\subsection{Explanation Similarity}
\label{subsec:explanation-comparison}

To better understand the underlying generation pattern of LLMs, as well as the similarities and differences between LLM and human reasoning, we analyze the linguistic similarity between human and LLM-generated explanations before and after validation.
Following \citet{chen2025rosenamellmgeneratedexplanations, hong2025litexlinguistictaxonomyexplanations}, we measure lexical, syntactic, and semantic similarity between explanations \citep{giulianelli-etal-2023-comes}, where lower scores indicate greater variation and the higher the more similar.
We compare pairwise similarity within-\varierr{} human explanations (\textit{within-human}), within each LLM (\textit{within-LLM}), and between each LLM and human explanations (\textit{LLM-vs-human}). For each item, we compute pairwise scores among all explanations under the same label and average across labels. 
Table \ref{tab:liguistic-similarity} reports results before validation and after three validation scenarios (\textit{one-expl}, \textit{one-llm}, and \textit{all-llm}); more results are in Appendix \ref{appx:similarity-full} Table \ref{tab:linguistic-similarity-full}.

Firstly, when comparing similarity scores between LLM sizes, larger models more frequently exhibit higher similarity in both \textit{within-LLM} and \textit{LLM-vs-human} regimes. 
\textit{Within-LLM} scores on Llama-7B explanations are an exception, scoring noticeably higher than Llama-70B on lexical and syntactic similarities.
Moreover, Llama-8B attains the highest scores across lexical, syntactic, and semantic dimensions among all LLMs in the \textit{within-LLM} regime, likely reflecting the greater volume and thus redundancy of its generated explanations.

Secondly, comparing \textit{within-human} and \textit{within-LLM} similarity scores, LLM generations exhibit greater lexical divergence (except for Llama-8B) while achieving higher syntactic and semantic similarity to each other than between human-written explanations, indicating that LLMs can generate lexically different outputs, while keeping syntactical structure and semantic meaning quite similar. 

Thirdly, the similarity between LLM- and human-generated explanations (\textit{LLM-vs-human}) is consistently lower than that observed \textit{within-human/LLM}, indicating that LLM generations still considerably diverge from human-written explanations, particularly in syntax and semantics.

Lastly, validation has minimal effects on similarity scores, typically showing a difference of less than 0.010 before versus after validation. Even so, most similarity scores (except for Llama-8B) increase after validation, reflecting the objective of removing low-quality explanations.

\subsection{Post-Validation Analyses} 
\label{subsec:after-validation}

\begin{table}[t]
\centering
\footnotesize
\resizebox{\linewidth}{!}{%
\begin{tabular}{l|lrrr|rrr}
\toprule
\textbf{Model}& \textbf{Validation} 
& \textbf{\# Expl.}
& \textbf{\# L/I} & \textbf{\# E/L} 
& AP & P@100 & R@100
\\
\midrule
\multirow{3}{*}{Llama-8B}  
& \textit{one-expl} & \textbf{7474} & \textbf{2.99} & \textbf{4.98} & \textbf{15.3}  &  \textbf{21.0} &   \textbf{16.3}\\
& \textit{one-llm} & 5951 & 2.32 & 3.97 & 13.9  &  12.0  &  9.3\\
& \textit{all-llm} & 6283 & 2.35 & 4.19 & 13.9  &  14.0 &  10.9 \\  
\midrule
\multirow{3}{*}{Llama-70B} &  \textit{one-expl} & \textbf{3354} & \textbf{2.82} & \textbf{2.24} & 18.4  &  24.0  &  18.6\\ 
& \textit{one-llm} & 2931 & 2.38 & 1.95 &  22.6  &  26.0 &   20.2\\
& \textit{all-llm} & 2799 & 2.33 & 1.87 & \textbf{26.8}  &  \textbf{31.0}  &  \textbf{24.0}\\
\midrule
\multirow{3}{*}{Qwen-7B} 
& \textit{one-expl} & \textbf{3282} & \textbf{2.73} & \textbf{2.19} & \textbf{16.4}  &  \textbf{19.0}  &  \textbf{15.3}\\ 
& \textit{one-llm} & 3157 & 2.44 & 2.10 & 13.6  &  15.0  &  12.1\\
& \textit{all-llm} & 1883 & 1.49 & 1.26 & 14.5  &  17.0  &  13.7\\
\midrule
\multirow{3}{*}{Qwen-72B} & \textit{one-expl} & \textbf{3286} & \textbf{2.83} & \textbf{2.19} & \textbf{18.8}  &  \textbf{26.0}  &  \textbf{20.2}\\ 
& \textit{one-llm} & 2878 & 2.45 & 1.92 & 16.7  &  21.0  &  16.3\\
& \textit{all-llm} & 2695 & 2.15 & 1.80 & \textbf{18.8}  &  22.0  &  17.1\\
\midrule
\multirow{2}{*}{VariErr} & self & 1712  & 1.50 & 2.29 & 22.8 & 23.7 & 18.3 \\
& random & 1712  & 1.50 & 2.29 & 14.7 & 14.7 & 11.4 
\\
\bottomrule
\end{tabular}
}
\caption{Statistics after validation and performance of \framework{} on the AED task. 
The AED values for \varierr{} with mean Datamaps and random baseline are copied from \citet{weber-genzel-etal-2024-varierr} for reference.}
\label{tab:validation_stats}
\end{table}

\paragraph{Statistics}
To assess the effectiveness of \framework{} for automatic error detection (AED), Table \ref{tab:validation_stats} presents the statistics after each validation scenario.
Compared with the pre-validation scores in Table \ref{tab:generation_stats}, 
LLM-based validation filters out a substantial proportion of generated explanations. Specifically, averaged across the three prompting strategies, Llama-8B excludes 25.7\% of explanations, Llama-70B 24.7\%, Qwen-7B 31.5\%, and Qwen-72B 24.7\%. Across all prompting strategies, \textit{all-llm} excludes the most generated explanations.
This further indicates that a validation strategy with a higher standard deviation effectively removes a larger proportion of invalid explanations.
LLMs exhibit a more radical filtering behavior than humans. Even under the mildest validation prompt (\textit{one-expl}), LLMs discard on average 16.8\% of the generated explanations, compared to 11.4\% for human validators.
Interestingly, the number of explanations per label in \varierr{} slightly increases after validation.
The removal of explanations more directly translates into the rejection of labels, likely due to the \varierr{}’s sparsity that many labels, especially erroneous ones, are supported by only a single explanation.

\paragraph{Automatic Error Detection (AED)} 
Following \citet{weber-genzel-etal-2024-varierr}, we also compute the average validation score per label for each instance and then use these scores to rank labels' likelihood of being erroneous. We evaluate this ranking against the self-flagged errors identified by human annotators.
We report standard ranking-based metrics: average precision (AP), as well as precision and recall at the top 100 predictions (P@100 and R@100) \cite{klie-etal-2023-annotation}.
We adopt the Datamaps model \citep{swayamdipta-etal-2020-dataset} and the random baseline from \varierr{} as our baselines.

Overall, larger models achieve stronger AED performance, while smaller models perform closely to the random baseline.
Among the validation strategies, \textit{one-expl} consistently yields the highest scores, while \textit{all-llm} with Llama-70B is the only configuration that surpasses the DM baseline.
Nevertheless, this evaluation setup has limitations. Since these metrics rely solely on the ranking of numerical scores, they may underrepresent labels that receive low scores but are not explicitly ranked lower.
For example, two instances with average scores of 0.75 and 0.05 would be classified as errors under a fixed threshold. 
We thus turn to fine-tuning experiments in \S\ref{sec:downstream-finetuning} using the analyzed thresholds (cf. Table~\ref{tab:validation_threshold}) that better preserve the numerical granularity.

\section{\framework{} for Fine-tuning} 
\label{sec:downstream-finetuning}

\renewcommand{\arraystretch}{1.2}
\begin{table}[t]
\centering
\footnotesize
\resizebox{\linewidth}{!}{%
\begin{tabular}{lc cc|cc}
\toprule
& & \multicolumn{2}{c}{\textbf{BERT}} & \multicolumn{2}{c}{\textbf{RoBERTa}} \\
\cmidrule(lr){3-4} \cmidrule(lr){5-6} 
\textbf{Models} & \textbf{Validation} & \multicolumn{1}{c}{\textbf{F1}~$\uparrow$} & \multicolumn{1}{c}{\textbf{KL}~$\downarrow$} 
& \multicolumn{1}{c}{\textbf{F1}~$\uparrow$} & \multicolumn{1}{c}{\textbf{KL}~$\downarrow$} \\[0.5ex]
\midrule
\multicolumn{2}{l}{\textbf{\varierr{} R1}  (\textit{baseline)}} & \textit{0.5842} & \textit{0.1701} &  \textit{0.6248} & \textit{0.1667} \\
\multicolumn{2}{l}{\textbf{\varierr{} R2} (\textit{baseline)}} &  0.5500 &   0.2510  & 0.6125 &   0.2474 \\
\hline
\multicolumn{6}{c}{\textit{(a) Fine-tuning with \framework{} labels}} \\
\multirow{4}{*}{\textbf{Llama-8B}}  & \textit{before} 
 & 0.4499 &  0.1181     & 0.3312     & 0.1188  \\
& \textit{one-expl}
&   0.4756 & 0.1172 & 0.5245  &  0.1170 \\
& \textit{one-llm}
& \textbf{0.5911} &     0.0971 &   0.5975 &   0.1218 \\
 & \textit{all-llm}
&  0.5820  &   \textbf{0.0891}    &  \textbf{0.6088}   &  \textbf{0.0984} \\
\hline
\multirow{4}{*}{\textbf{Llama-70B}} & \textit{before}   & 0.4499  & 0.1181  &   0.3312   & 0.1188  \\
& \textit{one-expl}    &  0.5219  &   0.1011 &   0.6119 &     0.0967 \\
&  \textit{one-llm}   & \textbf{0.6328}  & 0.0837 & 0.6151 &   0.0962 \\
&  \textit{all-llm}  & 0.6210  &   \textbf{0.0716} & \textbf{0.6351} &  \textbf{0.0818} \\
\hline
\multirow{4}{*}{\textbf{Qwen-7B}}  &  \textit{before} &  0.5074  &   0.1077 & 0.5406 & 0.1076  \\
&  \textit{one-expl} &   0.5519 &  0.1019 &   0.5624 &   0.1027 \\
 & \textit{one-llm} & \textbf{0.6294} &   \textbf{0.0783} &  \textbf{0.5983} & \textbf{0.0772}  \\
&  \textit{all-llm} &   0.4984  &  0.2566 &   0.5172  &    0.2724 \\
\hline
\multirow{4}{*}{Qwen-72B}  &  \textit{before}  &  0.4499  &   0.1181 &  0.3312   & 0.1188  \\
&  \textit{one-expl}  & 0.5310 &  0.1021 &  0.5570 &  \textbf{0.1001} \\
 & \textit{one-llm} &  0.5125   &  0.0994 &  0.5579 &   0.1122\\
& \textit{all-llm} &  \textbf{0.6245} &   \textbf{0.0807} &  \textbf{0.6545} & 0.1109  \\
\hline
\multicolumn{6}{c}{\textit{(b) Using \framework{} to validate \varierr{} R1 and then finetuning}} \\
\multirow{3}{*}{\textbf{Llama-8B}} & \textit{one-expl}
& \textbf{0.5939} & \textbf{0.1714} &0.6385 &  \textbf{0.1699} \\
 & \textit{one-llm}
& 0.6195 & 0.1938 &  \textbf{0.6617} & 0.1907 \\
& \textit{all-llm}
 & 0.5929 &  0.1916 &   0.6429 &   0.2042 \\
\hline
\multirow{3}{*}{\textbf{Llama-70B}} & \textit{one-expl} & 0.5968 & \textbf{0.1754} &  0.6383  & \textbf{0.1722} \\
 &  \textit{one-llm} & 0.5951 &  0.1924  & 0.6556 & 0.1903 \\
& \textit{all-llm}   & \textbf{0.6096} &   0.1915 & \textbf{0.6762} &  0.1805 \\
\hline
\multirow{3}{*}{\textbf{Qwen-7B}}  & \textit{one-expl}  &   0.5892  & 0.1755 &  0.6294 &  0.1787 \\
 &  \textit{one-llm}  &  0.6232 &  \textbf{0.1704}   & 0.6654 &  \textbf{0.1588} \\
&  \textit{all-llm}  &  \textbf{0.6415} &  0.2456 & \textbf{0.6731} &  0.2793  \\
\hline
\multirow{3}{*}{\textbf{Qwen-72B}}  &  \textit{one-expl}  & 0.5858 &   \textbf{0.1869} &  0.6370 &  \textbf{0.1833} \\
 & \textit{one-llm}  &  0.5993 &  0.1920 &   0.6376 &   0.1939 \\
&  \textit{all-llm} &  \textbf{0.6228} &  0.1954 &  \textbf{0.6614} &  0.2099 \\
\bottomrule
\end{tabular}
}
\caption{Results of directly fine-tuning with \framework{} labels or using \framework{} to validate \varierr{} R1 labels and then finetuning. }
\label{tab:finetuning}
\end{table}

To assess whether \framework{} benefits downstream fine-tuning experiments, we approximate human judgment distribution (HJD) in ChaosNLI \citep{nie-etal-2020-learn}, following \citet{chen-etal-2024-seeing}.

\paragraph{Setups}
We apply \framework{} in two setups: (a) directly using \framework{} labels before and after each validation scenario, and (b) using \framework{} validations to remove LLM-detected errors from \varierr{} Round 1 (R1) and then fine-tuning with pruned-\varierr{}.
This allows us to assess whether \framework{}-validation is directly useful to modeling and whether it can help refine human annotations.

\paragraph{Baselines \& Ceiling}
We consider two fine-tuning baselines: directly fine-tuning with \varierr{} R1 (Round 1) annotations, and fine-tuning after removing human-identified errors in \varierr{} R2 human validation.
The latter investigates whether removing self-identified errors from \varierr{} leads to improvements in HJD alignment.

\paragraph{Models}
We adopt two small pre-trained language models, \textbf{bert-base-uncased} \cite{devlin-etal-2019-bert} and \textbf{roberta-base} \cite{liu2019robertarobustlyoptimizedbert} as backbones following \citet{chen-etal-2024-seeing}.

\paragraph{Fine-tuning}
Models are first fine-tuned on the large single-label MNLI training set (392k examples, \citealt{williams-etal-2018-broad}) and validated on the matched development set (9.8k examples) to learn the general structure of the NLI task.
They are then fine-tuned on the label distributions derived from \framework{} validations.
For each instance, we construct the label distribution by assigning equal probability to each candidate E/N/C label that appears in the set, and normalizing the values accordingly.
To evaluate the performance, we use the 1,099 ChaosNLI instances \citep{nie-etal-2020-learn} that do not overlap with \varierr{}, and split them into development and test sets, containing 549 and 550 instances, respectively. We use the label distributions of 100 crowdworkers in ChaosNLI as the gold standard and evaluate model performance in predicting soft labels using KL divergence and weighted F1.
See Appendix \ref{appx:hyperparams} Tables \ref{tab:mnli-hyperparams}-\ref{tab:varierr-hyperparams} for hyper-parameters.

\paragraph{Results}
Table \ref{tab:finetuning} presents the results. 
When comparing two \varierr{} baselines, contrary to expectations, removing human-detected errors (R2) does not improve alignment with HJD.
For BERT and RoBERTa finetuning, both \framework{}-informed setups improve over the pre-validation results (\textit{before} for setup (a) and R1 for setup (b)) in terms of F1.
However, KL divergence is only moderately reduced when directly using \framework{} labels in setup (a). 
While setup (b) does not show improvements over baseline R1 in terms of KL divergence, suggesting that \framework{} does not fully optimize the label distribution, it nevertheless demonstrates substantial improvements over R2, highlighting the potential of LLMs in detecting errors.

When evaluating the \framework{} validation performance, particularly on the improved weighted F1, we observe that \textit{all-llm} delivers the best performance in both setups. This substantiates our hypothesis that LLM validation with additional labels and explanations as context benefits modeling. 
In contrast, \textit{one-expl} performs slightly worse, despite achieving the strongest results in ranking-based AED evaluation (as shown in Table~\ref{tab:validation_stats}).
We attribute this discrepancy to a mismatch between ranking-based evaluation and distribution-based supervision. While \textit{one-expl} produces well-ordered but overconfident scores that benefit ranking, it retains more low-quality explanations, leading to less reliable and poorly calibrated label distributions for the fine-tuning task.

As for model sizes, the larger Llama-70B has an advantage over Llama-8B, but surprisingly, Qwen-7B beats Qwen-72B in BERT and RoBERTa fine-tuning on both F1 and KL in the \framework{}-pruning setup (b).
However, Qwen-7B significantly degrades when fine-tuned using \textit{all-llm} validation in setup (a).
These outlying results echo Qwen-7B's noticeably lower number of explanations after \textit{all-llm} validation reported in Table \ref{tab:validation_stats}.
Above all, our findings suggest that LLMs can effectively help approximate HJD by validating their annotations using \framework{} or by validating human ones.

\section{Conclusion}

In this paper, we introduce \framework{}, an LLM-based explanation generation and validation framework designed to remove annotation errors from human label variation (HLV). 
Our findings show that LLMs produce more comprehensive explanations than human annotators. By analyzing explanation distributions before and after LLM-based validation, we demonstrate that LLM-generated explanations reliably signal valid instances of HLV, and that the validation process effectively refines these distributions to better capture human annotation variability. 
Larger context inputs and more capable models further yield more consistent and rational validation outcomes.
Moreover, the validation process exhibits a high degree of consensus with human judgments in identifying valid labels, likely attributable to the broader range of candidate explanations considered by LLMs. Finally, we show that removing LLM-detected errors from the dataset yields superior fine-tuning performance compared to removing human-detected errors, underscoring the practical value of LLM-based validation for enhancing dataset quality.
Overall, these analyses indicate that LLMs achieve strong performance on error detection tasks, often reaching levels comparable to human annotators, while additionally providing broader explanatory coverage and requiring less human expert involvement.

\section*{Limitations}

While our framework demonstrates that LLM-based explanation generation and validation can effectively detect annotation errors from HLV in NLI tasks, several limitations remain.
First, the current study is restricted to the NLI task and to the \varierr{} dataset. Datasets that simultaneously provide HLV references and identified errors are scarce, which limits us to fully assess the robustness and generalizability of the framework across other tasks and domains.
Second, \varierr{} incorporates a peer-validation setup, where annotators evaluate not only their own explanations but also those of others. This setting potentially provides richer information for error detection, but it is not explored in the present study. Incorporating peer-validation signals through LLMs remains an important direction for future work.
Lastly, our \framework{} framework to use LLMs to detect errors is exploratory.
LLMs tend to overgenerate explanations across labels, which makes it challenging to use the same evaluation setup for natural or synthetic errors. 
Although this behavior may preserve ambiguity and variation, it also risks inheriting social or cultural biases from LLMs, potentially marginalizing minority perspectives. Future work could address this by incorporating more diverse LLMs and retaining low-validaty annotations as ambiguous cases for further review.




\section*{Use of AI Assistants}
The authors acknowledge the use of ChatGPT for grammatical correction, to improve the coherence of the final manuscripts, and to assist with coding-related tasks.

\section*{Acknowledgments} 

We thank the members of the MaiNLP lab and reviewers for their valuable and constructive feedback. We are especially grateful to Beiduo Chen and Maribel Acosta for insightful suggestions on the draft of the paper. BP acknowledges funding by ERC Consolidator Grant DIALECT 101043235. LZ acknowledges support from the TUM-IAS Dieter Schwarz Fellowship.

\bibliography{acl_latex}

\appendix
\clearpage

\section{Prompt Templates}
\label{sec:appx-prompt}

This section presents the prompt templates used in this study for explanation generation (\S\ref{subsec:appx-generation-prompt}) and validation (\S\ref{subsec:appx-validation-prompt}).

\subsection{Explanation Generation Prompt}\label{subsec:appx-generation-prompt}

Figure \ref{fig:generate_explanation} shows the explanation generation prompt adapted from \citet{chen2025rosenamellmgeneratedexplanations}, with additional instructions allowing the LLM to abstain when a label cannot be reasonably justified. The prompt also explicitly discourages introductory phrases and semantic redundancy.

\begin{figure}[h]
\centering
\small
\fcolorbox{white}{gray!8}{
  \parbox{.95\linewidth}{
    \ttfamily
    \textbf{EXPLANATION GENERATION PROMPT} \\ [1ex]
    \textbf{"role": "system", "content":}  \\
            You are an expert in Natural Language Inference (NLI). List every distinct explanation for why the statement is \{relationship\} given the context below without introductory phrases. \\ If you think the relationship is false given the context, you can choose not to provide explanations. Do not repeat or paraphrase the same idea in different words. End your answer after all reasonable distinct explanations are listed. \\ Format your answer as a numbered list (e.g., 1., 2., 3.) \\
    \\
    "role": "user", "content":  \\
    Context: \{premise\} \\
    Statement: \{hypothesis\} \\ 
  }
}
\caption{Explanation generation prompt.}
\label{fig:generate_explanation}
\end{figure}

\subsection{Explanation Validation Prompts}\label{subsec:appx-validation-prompt}

Following the prompt design of \citet{weber-genzel-etal-2024-varierr} with with several modifications, Figure \ref{fig:validate-prompt-one-expl} presents the prompt for \textit{one-expl} setting
and Figure \ref{fig:validate-prompt-one-llm-all-llms} illustrates the prompt for \textit{one-llm} and \textit{all-llm} settings. 
The latter one explicitly instructs the model to utilize the contextual information provided and to generate all validity scores together in a structured JSON format.

\begin{figure}[h]
\centering
\small
\fcolorbox{white}{gray!8}{
  \parbox{.95\linewidth}{
    \ttfamily
    \textbf{ORIGINAL VALIDATION PROMPT} \\[1ex]
    You are an expert linguistic annotator. \\[0.5ex]
    We have collected annotations for an NLI instance together with reasons for the labels. 
    Your task is to judge whether the reasons make sense for the label. 
    Provide the probability (0.0--1.0) that the reason makes sense for the label. 
    Give ONLY the probability, no other words or explanation. \\[1ex]
    For example: \\
    Probability: <the probability between 0.0 and 1.0 that the reason makes sense for the label, without any extra commentary whatsoever; just the probability!> \\[1ex]
    Context: \{premise\} \\
    Statement: \{hypothesis\} \\
    Reason for label \{label\}: \{reason\_text()\} \\
    Probability:
    \\
}
}
\caption{Validation prompt for \textit{one-expl} scenario.}
\label{fig:validate-prompt-one-expl}
\end{figure}

\begin{figure}[H]
\centering
\small
\fcolorbox{white}{gray!8}{
  \parbox{.9\linewidth}{
    \ttfamily
    \textbf{ALL VALIDATION PROMPT} \\[1ex]
    You are an expert linguistic annotator. \\[0.5ex]
    We have collected annotations for an NLI instance together with explanations for the labels. 
    You will first be shown all explanations together so that you understand the overall context, and then your task is to judge whether each reason makes sense for the label. You must output a single JSON object that maps each explanation's index (1,2,3,...) to its probability in one time.  \\[0.5ex]
    Provide the probability (0.0 - 1.0) that each reason makes sense for the label. Give ONLY the probability, no other words or explanation. \\[1ex]
    Output example: \{``1'': 0.9, ``2'': 0.8, ...\} \\[1ex]
    Context: \{premise\} \\
    Statement: \{hypothesis\} \\
    Reason \{i\} for label \{label\}: \{reason\_text()\} \\
    Reason \{i\} for label \{label\}: \{reason\_text()\} \\
    ... \\[1ex]
    Now output the JSON object ONLY.
    \\
}
}
\caption{Validation prompt for \textit{one-llm} and \textit{all-llm} scenarios.}
\label{fig:validate-prompt-one-llm-all-llms}
\end{figure}

\section{Fine-Tuning Hyper-parameters}
\label{appx:hyperparams}

Hyper-parameter configurations for fine-tuning BERT and RoBERTa on the MNLI and \varierr{} dataset variants are summarized in Tables~\ref{tab:mnli-hyperparams} and~\ref{tab:varierr-hyperparams}, respectively.

\begin{table}[H]
\centering
\resizebox{0.8\linewidth}{!}{%
\begin{tabular}{ll}
\toprule
\textbf{Hyper-parameter} & \textbf{Value} \\ 
\midrule
Learning Rate Decay & Linear \\
Weight Decay        & 0.0 \\
Optimizer           & AdamW \\
Max sequence length & 128 \\
Learning Rate       & 2e-5 \\
Batch size          & 16 \\
Num Epoch           & 3 \\
Metric for best model  & eval\_accuracy \\
\bottomrule
\end{tabular}%
}
\caption{Hyper-parameters used for fine-tuning BERT and RoBERTa on the MNLI dataset.}
\label{tab:mnli-hyperparams}
\end{table}

\begin{table}[H]
\centering
\resizebox{0.8\linewidth}{!}{%
\begin{tabular}{ll}
\toprule
\textbf{Hyper-parameter} & \textbf{Value} \\
\midrule
Learning Rate Decay & Linear \\
Weight Decay        & 0.0 \\
Optimizer           & AdamW \\
Learning Rate       & 2e-5 \\
Batch size          & 4 \\
Num Epoch           & 5 \\
Metric for best model & eval\_macro\_F1 \\
\bottomrule
\end{tabular}%
}
\caption{Hyper-parameters used for further fine-tuning BERT and RoBERTa on the \varierr{} dataset variants.}
\label{tab:varierr-hyperparams}
\end{table}

\section{Validation Threshold}
\label{appx:validation-threshold}

Table~\ref{tab:validation_threshold} summarizes the optimal validation thresholds determined for each model and setting.
These thresholds correspond to the points that best balance lower KL divergence with higher precision and recall when comparing LLM-validated labels against the human-validated \varierr{} references, and are applied in all subsequent analysis.

\begin{table}[H]
\centering
\small
\begin{tabular}{lccc}
\toprule
 & \textit{one-expl} & \textit{one-llm} & \textit{all-llm}  \\
\midrule
\textbf{Llama-8B} & 0.8 & 0.2  & 0.2  \\
\textbf{Llama-70B} & 0.9 & 0.6 & 0.6 \\
\textbf{Qwen-7B} & 0.7 & 0.2  & 0.2 \\
\textbf{Qwen-72B} & 0.8 & 0.7  & 0.7\\
\bottomrule
\end{tabular}
\caption{Validation threshold.}
\label{tab:validation_threshold}
\end{table}

\section{Full Similarity Results}
\label{appx:similarity-full}

Table~\ref{tab:linguistic-similarity-full} reports full linguistic similarity results between human and LLM explanations across lexical ($n$-gram overlap), syntactic (POS $n$-gram overlap), and semantic (cosine and Euclidean) levels. For lexical and syntactic similarity, we compute $n$-gram overlaps with $n=1,2,3$.
Scores are shown for four stages: before validation, after \textit{one-expl} validation, after \textit{one-llm} validation, and after \textit{all-llm} validation.

\begin{table*}[t]
\centering
\renewcommand{\arraystretch}{1.15}
\resizebox{\linewidth}{!}{
\footnotesize
\begin{tabular}{llccc|ccc|cc|c}
\toprule
\multicolumn{1}{c}{\multirow{2}{*}{\textbf{Models}}} & \multicolumn{1}{c}{\multirow{2}{*}{\textbf{Setting}}} & \multicolumn{3}{c|}{\textbf{Lexical}} & \multicolumn{3}{c|}{\textbf{Syntactic}} & \multicolumn{2}{c|}{\textbf{Semantic}} & \textbf{AVG}\\
\cmidrule(lr){3-11}
& & {n = 1$\downarrow$} & {n = 2$\downarrow$} & {n = 3$\downarrow$} & {n = 1$\downarrow$} & {n = 2$\downarrow$} & {n = 3$\downarrow$} & {Cos.$\downarrow$} & {Euc.$\downarrow$} & {AVG $\downarrow$} \\
\midrule
within-human &  & 0.313 & 0.118 & 0.052 & 0.713 & 0.323 & 0.144 & 0.529 & 0.522 & 0.339 \\
\midrule
\multicolumn{11}{l}{\textit{within-LLM}} \\
\multirow{4}{*}{\, Llama-8B}
& before     & 0.382 & 0.191 & 0.103 & 0.841 & 0.483 & 0.265 & 0.599 & 0.542 & 0.426 \\
& one-expl   & 0.382 & 0.189 & 0.101 & 0.840 & 0.483 & 0.264 & 0.608 & 0.545 & 0.427 \\
& one-llm    & 0.378 & 0.183 & 0.097 & 0.842 & 0.484 & 0.263 & 0.611 & 0.545 & 0.425 \\
& all-llm    & 0.380 & 0.186 & 0.099 & 0.842 & 0.483 & 0.263 & 0.606 & 0.544 & 0.425 \\
\hline
\multirow{4}{*}{\, Llama-70B}
& before     & 0.310 & 0.100 & 0.030 & 0.841 & 0.444 & 0.201 & 0.595 & 0.535 & 0.382 \\
& one-expl   & 0.313 & 0.100 & 0.031 & 0.840 & 0.445 & 0.201 & 0.606 & 0.539 & 0.384 \\
& one-llm    & 0.308 & 0.097 & 0.029 & 0.838 & 0.446 & 0.200 & 0.605 & 0.538 & 0.383 \\
& all-llm    & 0.309 & 0.097 & 0.029 & 0.836 & 0.447 & 0.202 & 0.608 & 0.540 & 0.384 \\
\hline
\multirow{4}{*}{\, Qwen-7B}
& before     & 0.211 & 0.050 & 0.010 & 0.804 & 0.409 & 0.160 & 0.533 & 0.517 & 0.337 \\
& one-expl   & 0.219 & 0.053 & 0.012 & 0.806 & 0.411 & 0.161 & 0.548 & 0.521 & 0.341 \\
& one-llm    & 0.207 & 0.048 & 0.010 & 0.804 & 0.415 & 0.164 & 0.536 & 0.518 & 0.338 \\
& all-llm    & 0.200 & 0.042 & 0.010 & 0.805 & 0.428 & 0.170 & 0.536 & 0.518 & 0.339 \\
\hline
\multirow{4}{*}{\, Qwen-72B}
& before     & 0.266 & 0.083 & 0.023 & 0.824 & 0.431 & 0.188 & 0.578 & 0.530 & 0.365 \\
& one-expl   & 0.270 & 0.084 & 0.023 & 0.824 & 0.431 & 0.189 & 0.583 & 0.532 & 0.367 \\
& one-llm    & 0.265 & 0.079 & 0.021 & 0.824 & 0.432 & 0.187 & 0.583 & 0.531 & 0.365 \\
& all-llm    & 0.263 & 0.080 & 0.022 & 0.823 & 0.435 & 0.191 & 0.584 & 0.532 & 0.366 \\
\midrule
\multicolumn{11}{l}{\textit{LLM-vs-human}} \\
\multirow{4}{*}{\, Llama-8B}
& before     & 0.233 & 0.062 & 0.020 & 0.742 & 0.308 & 0.116 & 0.450 & 0.497 & 0.303 \\
& one-expl   & 0.234 & 0.062 & 0.020 & 0.739 & 0.307 & 0.116 & 0.452 & 0.498 & 0.303 \\
& one-llm    & 0.233 & 0.061 & 0.019 & 0.742 & 0.310 & 0.117 & 0.453 & 0.498 & 0.304 \\
& all-llm    & 0.233. &  0.060& 0.018 & 0.745 & 0.309 &0.116&  0.453 & 0.498 & 0.304 \\
\hline
\multirow{4}{*}{\, Llama-70B}
& before     & 0.257 & 0.079 & 0.028 & 0.754 & 0.322 & 0.129 & 0.515 & 0.514 & 0.325 \\
& one-expl   & 0.262 & 0.081 & 0.029 & 0.755 & 0.325 & 0.131 & 0.521 & 0.516 & 0.328 \\
& one-llm    & 0.262 & 0.082 & 0.029 & 0.756 & 0.326 & 0.132 & 0.521 & 0.516 & 0.328 \\
& all-llm    & 0.263 & 0.083 & 0.030 & 0.753 & 0.326 & 0.132 & 0.523 & 0.517 & 0.328\\
\hline
\multirow{4}{*}{\, Qwen-7B}
& before     & 0.207 & 0.052 & 0.015 & 0.727 & 0.293 & 0.101 & 0.469 & 0.502 & 0.296 \\
& one-expl   & 0.213 & 0.054 & 0.016 & 0.728 & 0.296 & 0.103 & 0.477 & 0.504 & 0.299 \\
& one-llm    & 0.206 &  0.052 & 0.015& 0.725 & 0.293 & 0.100 & 0.469 & 0.502 & 0.295 \\
& all-llm    & 0.199 & 0.048 & 0.014 & 0.716 & 0.288 & 0.095 & 0.470 &  0.502 & 0.291 \\
\hline
\multirow{4}{*}{\, Qwen-72B}
& before     & 0.245 & 0.072 & 0.023 & 0.749 & 0.323 & 0.125 & 0.502 & 0.511 & 0.319 \\
& one-expl   & 0.248 & 0.074 & 0.024 & 0.748 & 0.324 & 0.127 & 0.506 & 0.512 & 0.320 \\
& one-llm    & 0.247 & 0.073 & 0.023 & 0.747 & 0.324 & 0.126 & 0.507 & 0.512 & 0.320 \\
& all-llm    & 0.247 & 0.073 & 0.024 & 0.745 & 0.324 & 0.126 & 0.504 & 0.512 & 0.319 \\
\bottomrule
\end{tabular}}
\caption{Full results comparing human and LLM explanations across lexical, syntactic, and semantic similarity. Each cell lists scores in the order: before validation, after \textit{one-expl} validation, after \textit{one-llm} validation, and after \textit{all-llm} validation.}
\label{tab:linguistic-similarity-full}
\end{table*}

\end{document}